\pdfoutput=1
\documentclass[10pt,logo,copyright]{nvidiatechreport}
\makeatletter
\renewcommand{\abscontent}{%
  \noindent{\large\bfseries\headingfont Abstract}\par\vspace{0.5em}%
  {\absfont \theabstract\par}%
  \@ifundefined{@keywords}{}{%
    \vskip1em \noindent \keywordsfont Keywords: \@keywords}%
}
\makeatother

\usepackage{algorithm}
\usepackage{algpseudocode}

\usepackage{adjustbox}

\usepackage[authoryear,sort&compress,round]{natbib}

\usepackage[utf8]{inputenc} % allow utf-8 input
\usepackage[T1]{fontenc}    % use 8-bit T1 fonts

\usepackage[printonlyused,nohyperlinks]{acronym} % acronym
\usepackage{soul, color}    % \hl{} support
\usepackage{parskip}        % no paragraph indents
\usepackage{url}            % simple URL typesetting
\usepackage{hyperref}       % hyperlinks
\usepackage{booktabs}       % professional-quality tables
\usepackage{amsfonts}       % blackboard math symbols
\usepackage{titlesec}
\usepackage{nicefrac}       % compact symbols for 1/2, etc.
\usepackage{microtype}      % microtypography
\usepackage{xcolor}         % colors
\usepackage[dvipsnames]{xcolor} % more color names
\usepackage{graphicx}
\usepackage{animate}        % for 360 video in teaser
\usepackage{comment}
\usepackage{subcaption}
\usepackage{tabularx}
\usepackage{makecell}
\usepackage{adjustbox}
\usepackage{setspace}
\newcolumntype{M}[1]{>{\centering\arraybackslash}m{#1}}
\usepackage{float}
\usepackage{tikz}
\usetikzlibrary{positioning,shapes,arrows}
\usepackage{amsmath,amsfonts,bm, bbm,leftindex}
\usepackage{multirow}
\usepackage{multicol}
\usepackage{comment}
\usepackage{gensymb}
\usepackage{lipsum}
\usetikzlibrary{arrows.meta, positioning, fit}
\usepackage[para]{threeparttable}
\usepackage{tikz}
\usetikzlibrary{tikzmark}

\usepackage{CJKutf8}
\usepackage[stable]{footmisc}

\usepackage{overpic}

\definecolor{backcolour}{rgb}{0.95,0.95,0.92}

\def\eqref#1{equation~\ref{#1}}
\def\1{\bm{1}}

\DeclareMathAlphabet{\mathsfit}{\encodingdefault}{\sfdefault}{m}{sl}
\SetMathAlphabet{\mathsfit}{bold}{\encodingdefault}{\sfdefault}{bx}{n}
\makeatletter
\let\save@mathaccent\mathaccent
\newcommand*\if@single[3]{%
  \setbox0\hbox{${\mathaccent"0362{#1}}^H$}%
  \setbox2\hbox{${\mathaccent"0362{\kern0pt#1}}^H$}%
  \ifdim\ht0=\ht2 #3\else #2\fi
  }
\newcommand*\rel@kern[1]{\kern#1\dimexpr\macc@kerna}
\newcommand*\widebar[1]{\@ifnextchar^{{\wide@bar{#1}{0}}}{\wide@bar{#1}{1}}}
\newcommand*\wide@bar[2]{\if@single{#1}{\wide@bar@{#1}{#2}{1}}{\wide@bar@{#1}{#2}{2}}}
\newcommand*\wide@bar@[3]{%
  \begingroup
  \def\mathaccent##1##2{%
    \let\mathaccent\save@mathaccent
    \if#32 \let\macc@nucleus\first@char \fi
    \setbox\z@\hbox{$\macc@style{\macc@nucleus}_{}$}%
    \setbox\tw@\hbox{$\macc@style{\macc@nucleus}{}_{}$}%
    \dimen@\wd\tw@
    \advance\dimen@-\wd\z@
    \divide\dimen@ 3
    \@tempdima\wd\tw@
    \advance\@tempdima-\scriptspace
    \divide\@tempdima 10
    \advance\dimen@-\@tempdima
    \ifdim\dimen@>\z@ \dimen@0pt\fi
    \rel@kern{0.6}\kern-\dimen@
    \if#31
      \overline{\rel@kern{-0.6}\kern\dimen@\macc@nucleus\rel@kern{0.4}\kern\dimen@}%
      \advance\dimen@0.4\dimexpr\macc@kerna
      \let\final@kern#2%
      \ifdim\dimen@<\z@ \let\final@kern1\fi
      \if\final@kern1 \kern-\dimen@\fi
    \else
      \overline{\rel@kern{-0.6}\kern\dimen@#1}%
    \fi
  }%
  \macc@depth\@ne
  \let\math@bgroup\@empty \let\math@egroup\macc@set@skewchar
  \mathsurround\z@ \frozen@everymath{\mathgroup\macc@group\relax}%
  \macc@set@skewchar\relax
  \let\mathaccentV\macc@nested@a
  \if#31
    \macc@nested@a\relax111{#1}%
  \else
    \def\gobble@till@marker##1\endmarker{}%
    \futurelet\first@char\gobble@till@marker#1\endmarker
    \ifcat\noexpand\first@char A\else
      \def\first@char{}%
    \fi
    \macc@nested@a\relax111{\first@char}%
  \fi
  \endgroup
}
\makeatother
\usepackage{algorithm}
\usepackage{algpseudocode}
\usepackage{subcaption}

\usepackage{xcolor} % for colors

\newcommand{\sectionowner}[2][]{%
  \textcolor{red}{
    \textbf{Section Owner:} #2%
    \ifthenelse{\equal{#1}{}}{}{ \textit{(#1)}}%
  }
}

\newcommand{\todo}[2][]{%
  \textcolor{blue}{
    \textbf{TODO:} #2%
    \ifthenelse{\equal{#1}{}}{}{ \textit{(#1)}}%
  }
}

\usepackage{pifont}
\newcommand{\cmark}{\ding{51}} % Tick mark
\newcommand{\xmark}{\ding{55}} % Cross mark
\usepackage{wrapfig}

\usepackage[nameinlink]{cleveref}
\crefname{equation}{Eq.}{Eqs.}
\crefname{figure}{Figure}{Figures}
\crefname{section}{Section}{Sections}
\crefname{appendix}{App.}{App.}
\crefname{table}{Table}{Tables}
\crefname{algorithm}{Algorithm}{Algorithms}
\crefname{thm}{Thm}{Thm}
\Crefname{thm}{Thm}{Thm}
\crefname{prop}{Prop}{Prop}
\usepackage{longtable}

\newcommand{\Sysname}{\textsc{EgoHand}\xspace}
\newcommand{\sysname}{\textsc{EgoHand}\xspace}

\DeclareMathDelimiter{(}{\mathopen} {operators}{"28}{largesymbols}{"00}
\DeclareMathDelimiter{)}{\mathclose}{operators}{"29}{largesymbols}{"01}

\newcolumntype{C}{>{\centering\arraybackslash}X}

\title{\Sysname: Monocular 3D Hand Reconstruction from Egocentric Video}

\author{NVIDIA\footnote{A detailed list of contributors and acknowledgments can be found in~\cref{sec:contributors} of this paper.}}

\date{October 2026}

\begin{abstract}

We introduce \Sysname, a monocular 3D hand reconstruction model that achieves state-of-the-art performance on egocentric video. Central to our approach is the curation of a high-quality training corpus of 14 million images, combining synthetic data, in-the-wild video, and multi-camera motion capture to balance scene diversity and supervision quality. The model adopts SOMA-Hand as its parametric representation and processes 2D hand crops with a transformer encoder, incorporating camera cues and an auxiliary 2D branch to improve geometric consistency and image alignment. Zero-shot evaluations on HOT3D, ARCTIC, and H2O demonstrate strong out-of-domain generalization, with lower joint-position, root-translation, and acceleration errors than competing baselines at an inference latency of approximately 9.01 ms. A blinded user preference study further shows that participants favor \Sysname's reconstructed hand poses over those from Apple EgoDex and Ropedia Xperience-10M. We release the \Sysname model and source code under the OpenMDW license, allowing free commercial use of the model.

Model:
\url{https://huggingface.co/nvidia/EgoHand-v1}

Inference:
\url{https://github.com/NVIDIA/IsaacCapture/tree/main/src/python/isaaccapture_egopost}

Training:
\url{https://github.com/NVlabs/GEM-X} % TODO: the folder will be updated to include EgoHand training

\end{abstract}

\begin{document}

\maketitle 

\abscontent

\vspace{5mm}

\begin{center}
    \includegraphics[width=\linewidth]{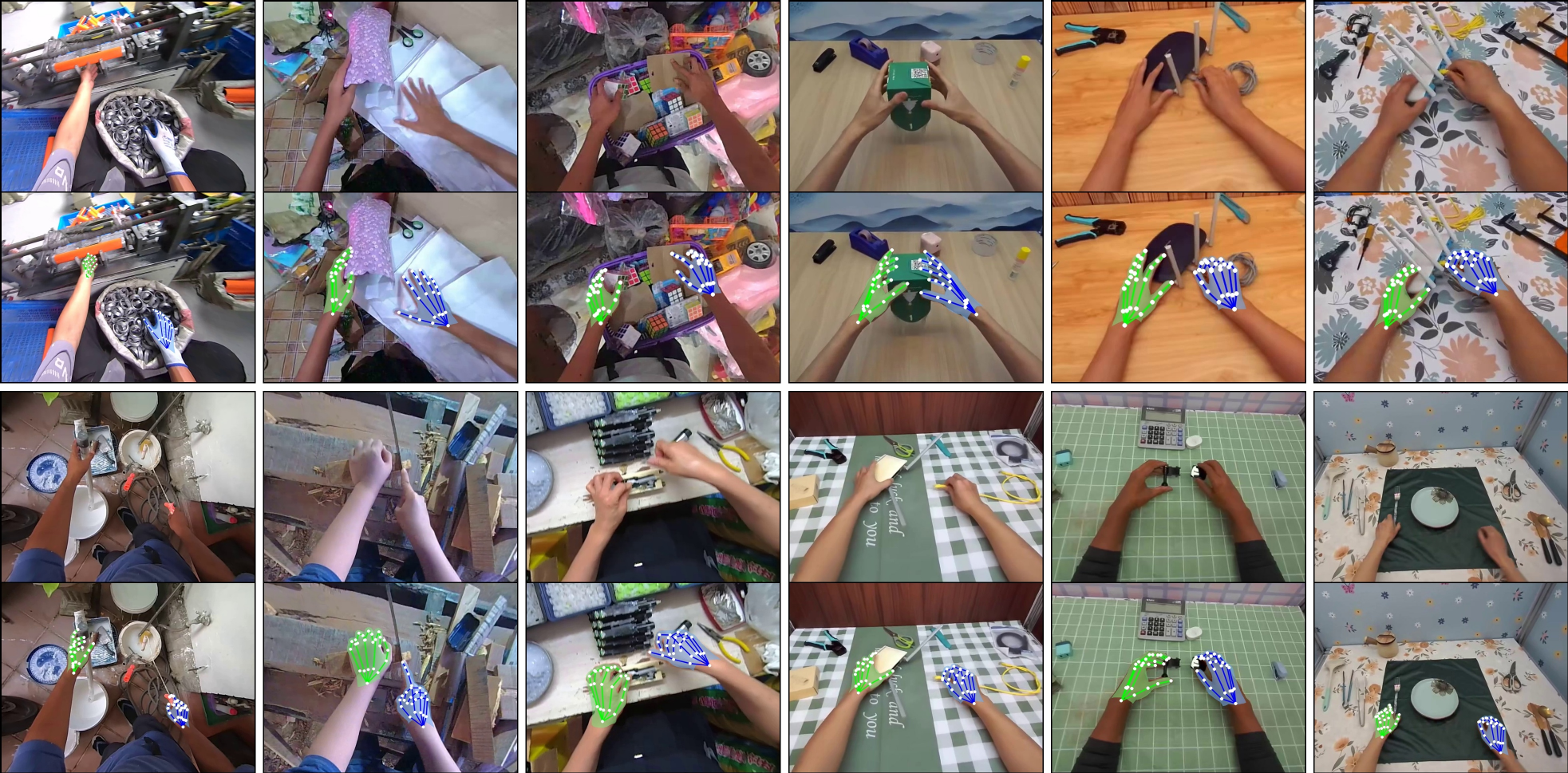}
    \captionof{figure}{\textbf{\sysname} hand reconstruction results. Our approach achieves robust reconstructions under challenging poses across diverse environments.
    % \ui{add more rows of qualitative results to complete the page}
    }
    \label{fig:teaser}
\end{center}
\clearpage

\section{Introduction}
\label{sec:intro}

Egocentric human videos paired with 3D hand annotations have emerged as a scalable source of supervision for pretraining vision-language-action (VLA) and world-action models (WAMs)~\citep{egoscale}. By capturing human manipulation across diverse objects, tasks, and environments, these recordings offer a path to reducing reliance on costly robot teleoperation. EgoScale~\citep{egoscale}, for example, demonstrates that scaling action-labeled human video improves dexterous manipulation performance and, together with aligned human--robot training, enables adaptation with few robot demonstrations. In these pipelines, reconstructed hand trajectories provide action supervision that connects visual observations to manipulation behavior. Accurate and efficient 3D hand reconstruction is therefore an important enabling capability for scaling robot learning from human demonstrations.

Despite substantial progress in monocular hand reconstruction~\citep{pavlakos2024hamer,potamias2025wilor,zhang2025hawor,yu2025dynhamr,wang2026vidihand}, producing reliable annotations across diverse egocentric recordings remains challenging. Models must generalize across hand appearances, backgrounds, objects, and camera configurations, yet obtaining accurate 3D supervision with comparable diversity is difficult. Combining heterogeneous datasets further introduces differences in joint definitions, annotation reliability, and geometric consistency. At inference time, reconstruction must also preserve image alignment and recover camera-relative hand placement from cropped observations, while remaining efficient enough to process large video collections.

We introduce \Sysname, a monocular 3D hand reconstruction model that addresses these challenges through camera-aware prediction and the systematic curation of diverse, high-quality training data. Our approach combines a geometrically grounded reconstruction architecture with a corpus of 14 million images that brings together complementary forms of visual diversity and 3D supervision.

\textbf{Model.}
We adopt SOMA-Hand~\citep{saito2026soma} with the MHR shape backend~\citep{ferguson2025mhr}, which provides explicit skeletal scale controls alongside surface-shape parameters. This representation supports fitting heterogeneous landmark annotations while maintaining a consistent hand identity across each sequence. Following prior methods~\citep{pavlakos2024hamer,potamias2025wilor}, \sysname processes 2D hand crops with a transformer encoder. Because cropping and resizing remove explicit information about the hand's original image location and extent, we combine bounding-box conditioning~\citep{li2022cliff} with ray positional encoding~\citep{prakash2024wildhands} to retain the geometric relationship between the crop and the camera. A separate 2D joint prediction branch supplies image-space correspondences that refine lateral wrist translation while retaining the depth predicted by the 3D branch. Joint training through this refinement couples image alignment with reconstructed geometry and camera-relative placement. Section~\ref{sec:method} describes the model.

\textbf{Data.}
We construct our training corpus from three complementary sources. First, we use high-quality hand motion captured with gloves to animate synthetic hands with diverse shapes and appearances. This provides abundant training images with accurate geometric labels and controlled variation in hand appearance. Second, we incorporate in-the-wild egocentric videos to capture the breadth of backgrounds, objects, and interactions needed for robust generalization. To make these recordings suitable for training, we develop a comprehensive pseudo-annotation pipeline that combines available 3D landmarks, image-derived hand segmentation masks, and temporal constraints. The pipeline estimates a shared hand shape within each sequence, optimizes per-frame articulation and wrist motion, and refines difficult intervals using alignment and motion checks. Third, we collect hand motion using micro-markers synchronized with multiview video, yielding real images paired with high-quality 3D ground-truth annotations. This source complements the appearance diversity of synthetic data and the scene and object diversity of in-the-wild video with precise supervision on real imagery, helping bridge the synthetic-to-real domain gap. We express all annotations in a common SOMA-Hand representation to support consistent training across sources. Section~\ref{sec:dataset} details the data collection and annotation pipeline.

\sysname achieves state-of-the-art 3D hand reconstruction with an inference
latency of approximately 9.01 ms on an NVIDIA RTX Pro 6000 GPU.
It demonstrates strong zero-shot generalization on HOT3D~\citep{banerjee2025hot3d}, H2O~\citep{kwon2021h2o}, and ARCTIC~\citep{fan2023arctic} without training on any of these datasets. On H2O, it achieves the lowest mean and p90 joint-position, root-translation, and acceleration errors among HaMeR~\citep{pavlakos2024hamer}, WiLoR~\citep{potamias2025wilor}, HaWoR~\citep{zhang2025hawor}, and our best-effort reproduction of Dyn-HaMR~\citep{yu2025dynhamr}. On HOT3D and ARCTIC, which are part of the training set of WiLoR and HaWoR, \sysname outperforms HaMeR and our Dyn-HaMR reproduction across all five evaluated metrics in zero-shot.
% EgoHand combines this accuracy with an inference latency of 18.62 ms per frame on an NVIDIA L40 GPU, or 20.18 ms including temporal smoothing. 
A blinded side-by-side study further shows that human graders prefer \sysname’s projected hand poses over the supplied annotations on 77.0\% of Apple EgoDex clips and 95.5\% of Ropedia Xperience-10M clips, while reporting comparable visual quality with Lightwheel EgoStandard. We release the \Sysname model and source code under the OpenMDW license, allowing commercial use of the model.

% To summarize, our contributions are as follows:
% \begin{itemize}
% \item First commercial friendly 3D hand reconstruction model achieves SOTA
% performance, with direct
% SOMA-Hand prediction to recover hand pose and shape
% sequences from monocular egocentric video.

% \item A data strategy that combines synthetic, in-the-wild, and
% multi-camera motion capture data to balance scene
% diversity with geometric supervision quality and a
% format conversion algorithm to convert them into consistent SOMA-Hand.

% \item A model design that combines camera geometry cues, a separate
% 2D joint branch, and joint 2D and 3D supervision, with calibrated perspective
% cropping to support pinhole and fisheye cameras.

% \end{itemize}

\section{Related Work}
\label{sec:background}

\textbf{3D Hand Representations.}
MANO is widely used for hand pose and shape reconstruction~\citep{romero2017mano}.
Its learned shape space couples skeletal proportions and surface geometry,
limiting direct control of individual bone lengths.
Momentum Human Rig (MHR) decouples skeletal parameters from surface shape and
includes 23 joints per hand, counting the wrist and fingertip end
joints~\citep{ferguson2025mhr}. SOMA accommodates different skeletons and shape
backends in a common framework~\citep{saito2026soma}. For both annotation and
prediction, we use native SOMA-Hand with 25 joints per hand and the MHR shape
backend with explicit skeletal scale controls~\citep{nvidia2026somax}.

\textbf{Hand Reconstruction Models.}
HaMeR~\citep{pavlakos2024hamer} reconstructs hands using a large vision transformer,
while WiLoR~\citep{potamias2025wilor} combines hand localization and reconstruction.
For video, Dyn-HaMR~\citep{yu2025dynhamr} recovers world-coordinate hand trajectories through motion
priors and multi-stage optimization; HaWoR~\citep{zhang2025hawor} combines temporal
reconstruction, camera tracking, and motion infilling.
ViDiHand~\citep{wang2026vidihand} draws on scene and motion priors from a pretrained video diffusion model
to reconstruct heavily occluded hands. Its reported
throughput of 5.5 frames per second on four A100 GPUs limits its practicality for
large-scale annotation.
\sysname combines camera conditioning and a separate 2D joint branch to
constrain camera-space hand placement, using a DINOv3 encoder initialized from
SAM 3D Body~\citep{simeoni2025dinov3,yang2026sam3dbody}.
Compared with the evaluated baselines, it improves joint-position and
root-translation accuracy while retaining efficient single-frame inference.
% \tc{reemphasizing why we are better on perf, see SAM3D Body related work as a ref} -> updated 

\textbf{Hand Reconstruction Datasets.}
Hand reconstruction across varied poses and imaging conditions benefits from
large, diverse training sets~\citep{pavlakos2024hamer,potamias2025wilor}.
Existing resources span real captures, including ARCTIC and
HOT3D~\citep{fan2023arctic,banerjee2025hot3d}, and synthetic renderings such as
RHD and BEDLAM~\citep{zimmermann2017rhd,black2023bedlam}.
FreiHAND and WiLoR fit parametric hand models to skeletal keypoints to obtain
pose and shape annotations~\citep{zimmermann2019freihand,potamias2025wilor}.
To scale training with diverse appearances and 3D supervision, we combine
synthetic images, in-the-wild video, and multi-camera motion capture, totaling
roughly 14 million images. Synthetic data retain SOMA-Hand parameters; real
recordings are fitted to 3D landmarks in the same representation
(Section~\ref{sec:dataset}).
% \tc{Emphasize large-scale datasets are important, thus we use 3 sources, also add skeleton based fitting as prior work. shrink the details of each datasets} -> updated
\section{\sysname}
\label{sec:method}

% \subsection{System Overview}

% Given an egocentric 2D hand crop, \sysname reconstructs the full 3D hand pose, shape, and camera-space translation. The model architecture is illustrated in Figure~\ref{fig:egohand_architecture}.

\subsection{Network Architecture}

\begin{figure}[htbp]
\centering
\includegraphics[width=\linewidth]{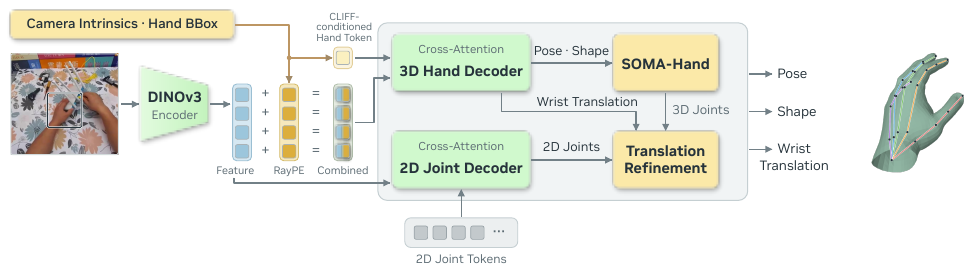}
\caption{\sysname model architecture. A DINOv3 encoder supports two branches. The 3D hand reconstruction branch uses a CLIFF-conditioned hand query to cross-attend to image features augmented with ray positional encoding, predicting SOMA-Hand pose, shape, and wrist translation. The 2D joint prediction branch uses separate learned joint queries to cross-attend to image features. Its predicted 2D joints, together with the SOMA-Hand 3D joints and camera intrinsics, refine lateral wrist translation while preserving the predicted depth, yielding hand geometry in camera coordinates.
% \ui{Make the caption detailed. I should be able to understand everything just from the caption. }
}
\label{fig:egohand_architecture}
\end{figure}

Given a 2D hand crop, \sysname extracts image features using a DINOv3
ViT-H/16+ encoder~\citep{simeoni2025dinov3}. As shown in Figure~\ref{fig:egohand_architecture}, the \textbf{3D hand reconstruction branch} (Section~\ref{sec:camera_awareprediction}) combines these features with geometric cues derived from camera intrinsics and the 2D bounding box to predict hand pose, shape, and wrist translation. A separate \textbf{2D joint prediction branch} (Section~\ref{sec:2Dbranch}) uses the image features to predict 2D joint locations. Translation refinement uses correspondences between predicted 2D and reconstructed 3D joints to adjust translation, positioning the reconstructed hand in camera coordinates. Together, the two branches recover hand geometry and its wrist translation in camera coordinates.

\subsection{3D Hand Reconstruction}
\label{sec:camera_awareprediction}

Resizing a hand crop standardizes the network input but removes explicit information about its position and extent in the original image. We retain this geometric context through bounding-box conditioning (CLIFF)~\citep{li2022cliff}, which encodes the crop's location and scale relative to the camera, and ray positional encoding (RayPE)~\citep{wang2026vidihand}, which associates image features with their viewing directions. %\tc{more succint, maybe merge with previous sentence} Bounding-box conditioning describes the crop's location and scale relative to the camera, while ray positional encoding associates individual image features with their viewing directions.
As illustrated in Figure~\ref{fig:egohand_architecture}, bounding-box conditions the initial decoder token, while ray features are added to the image features. From the 3D hand decoder output, regression heads predict SOMA-Hand pose and shape parameters. The pose parameters describe wrist orientation and finger articulation, while the shape parameters control hand shape and proportions. SOMA-Hand maps these predictions to a hand mesh and 3D joints expressed relative to the wrist. A separate camera head predicts the translation that positions this geometry in camera coordinates.

For a square crop with center $(c_x,c_y)$ and side length $b$, let
$(f_x,f_y)$ and $(p_x,p_y)$ denote the camera focal lengths and principal
point, respectively. The camera head predicts $(\ell,d_x,d_y)$, where $\ell=\log(bZ/f_x)$ parameterizes wrist depth $Z$, and $d_x,d_y$ are metric lateral offsets. These parameters yield a direct estimate of wrist translation in camera coordinates:
\begin{equation}
\begin{aligned}
Z &= \frac{f_x}{b}\exp(\ell),\\
T_x^{\mathrm{reg}} &= d_x+\frac{c_x-p_x}{f_x}Z,\\
T_y^{\mathrm{reg}} &= d_y+\frac{c_y-p_y}{f_y}Z.
\end{aligned}
\end{equation}

%The estimate $T^{\mathrm{reg}}=(T_x^{\mathrm{reg}},T_y^{\mathrm{reg}},Z)$ is obtained from the 3D hand reconstruction branch alone. %The correspondence solve in Section~\ref{sec:2Dbranch} combines predictions from both branches to estimate lateral translation using the full intrinsic matrix. When valid, this solution replaces the lateral components of $T^{\mathrm{reg}}$; otherwise, we retain the direct estimate. In both cases, the predicted depth $Z$ is unchanged.

% Given a reconstructed wrist-relative joint $J_j$ and the selected wrist translation $T=(T_x,T_y,Z)$, its full-image and crop-normalized projections are

% \begin{equation}
% q_j = \pi_K(J_j+T), \qquad
% \widetilde q_j = \frac{q_j-(c_x,c_y)}{b}.
% \end{equation}

% The projection $\pi_K$ uses each joint's camera-space depth.

\subsection{2D Joint Prediction and Translation Refinement}
\label{sec:2Dbranch}

The 2D joint prediction branch estimates joint locations from learnable 2D joint tokens by cross-attenting to the image features, providing correspondences between image observations and the hand geometry reconstructed by the 3D branch. These correspondences constrain lateral translation while retaining the wrist depth predicted by the 3D branch.

Let $\widehat q_j$ denote a predicted 2D joint mapped from crop-normalized coordinates to image pixels, and $J_j$ the corresponding reconstructed wrist-relative 3D joint. Given camera intrinsics $K$ and predicted wrist depth $Z$, we estimate lateral translation by minimizing reprojection error over selected correspondences $\mathcal V$:

\begin{equation}
\begin{aligned}
(\widehat T_x,\widehat T_y)
&=\underset{(T_x,T_y)\in\mathbb R^2}{\operatorname{argmin}}\;
\sum_{j\in\mathcal V}\left\|
\pi_K\!\left(J_j+(T_x,T_y,Z)\right)-\widehat q_j
\right\|_2^2+\rho(T_x^2+T_y^2).
\end{aligned}
\end{equation}

%With $Z$ fixed, each joint's projection denominator is constant with respect to $T_x,T_y$, yielding a linear least-squares problem. 
The projection $\pi_K$ uses each joint's camera-space depth. The coefficient $\rho$ controls a small regularization term for numerical stability. When the solve is valid, its solution replaces the lateral components of the direct translation estimate $T^{\mathrm{reg}}$ from Section~\ref{sec:camera_awareprediction}, giving $T=(\widehat T_x,\widehat T_y,Z)$. Otherwise, we retain $T^{\mathrm{reg}}$. %Both estimates share the same predicted depth.

The 2D decoder also predicts uncertainty scales and a spatial evidence map for each joint. The uncertainty scales characterize horizontal and vertical localization uncertainty, while the normalized evidence map receives spatial localization supervision. Its total evidence determines a presence score that estimates whether the joint lies within the valid crop, regardless of occlusion. These auxiliary outputs support the training objectives in Section~\ref{sec:training_loss}, and presence scores help select correspondences at inference.

%assigns a nonnegative value to each location on the 16 \texttimes{} 16 image-feature grid. Its normalized distribution receives spatial localization supervision, while its total evidence is mapped to a joint-presence score that estimates whether the joint lies within the valid crop, including when it is occluded. \cref{sec:training_loss} defines the corresponding training losses.

%At inference, a presence score derived from each joint's spatial evidence map, together with crop and image bounds, determines its eligibility for the solve. This score represents inclusion within the valid crop, including for occluded joints. During training, eligibility is determined from annotation validity and geometry instead of predicted presence or uncertainty; the solve itself still uses predicted 2D and 3D coordinates. Its differentiability allows translation supervision to jointly train the 2D predictions, reconstructed hand geometry, and depth. Appendix~\ref{sec:implementation_details} specifies the correspondence-selection and numerical-validity checks.

\subsection{Training Objective}
\label{sec:training_loss}

We jointly train the model using a weighted combination of hand reconstruction, direct 2D prediction, and translation losses, $\mathcal L=\sum_i\lambda_i\mathcal L_i$.

\textbf{Hand reconstruction.} We apply annotation-confidence-weighted $L_1$ losses to projected 2D joints and wrist-relative 3D joint positions, and squared-error losses to SOMA-Hand pose and shape parameters. The camera head receives smooth $L_1$ supervision on the standardized logarithmic-depth residual and squared-error supervision on lateral offsets. Together, these losses constrain hand geometry and image alignment while training the direct translation estimate independently of the correspondence solve.

\textbf{Direct 2D joint prediction.} We supervise predicted joint coordinates using residual log-likelihood estimation with learned coordinate uncertainty~\citep{li2021rle}, including labeled joints outside the crop. Binary cross-entropy trains presence scores to indicate whether annotated joints lie within the valid crop. For joints inside this region, a spatial loss supervises the normalized evidence maps using the continuous ranked probability score on their horizontal and vertical marginals.

\textbf{Translation supervision.} We apply squared-error supervision to the translation obtained from solving valid 2D--3D correspondence, using ground-truth camera-space translation as the target. Backpropagation through the solve couples the predicted 2D coordinates, reconstructed 3D geometry, and depth. The solved translation also enters the 2D reprojection loss. When the solve is invalid, reprojection uses the direct estimate $T^{\mathrm{reg}}$, and the sample is excluded from the translation loss.

\section{Data Sourcing and Annotation}
\label{sec:dataset}

Robust 3D hand reconstruction relies on large-scale datasets with accurate,
consistent annotations~\citep{pavlakos2024hamer,potamias2025wilor}. Yet commercially usable datasets with accurate 3D hand annotations and diverse scenes, poses, and hand shapes remain scarce.

\Sysname is trained on 14 million images drawn from three complementary sources — 
synthetic data, in-the-wild video, and multi-view motion capture (MoCap) 
— to balance scene and camera diversity with annotation accuracy. 
Our \textit{annotation-fitting pipeline} brings these sources into a common SOMA-Hand 
representation by leveraging its decoupled shape and pose parameterization: a 
single hand shape is optimized per clip, while per-frame poses are fit jointly 
to 2D hand segmentation masks and available 3D keypoint annotations.

\subsection{Data Sources}

\textbf{Synthetic data.} Synthetic images provide accurate hand geometry
and controlled variation in camera configurations. We use NVIDIA
Omniverse~\citep{nvidia_omniverse} to generate approximately 49 hours of synthetic hand
sequences, varying hand size, motion, scene content, camera viewpoint, and
intrinsic parameters. 

\textbf{In-the-wild video.} In-the-wild footage contributes diverse appearances, 
scenes, object interactions, and hand sizes. We source around 63 hours licensed 
video, covering everyday manipulation
and industrial tasks, including pick-and-place, sorting, and the
assembly and disassembly of small rigid parts. The recordings use
head-mounted cameras with horizontal FOV ranging from $101^\circ$ to $165^\circ$, encompassing both wide-angle pinhole and fisheye camera models.

\textbf{MoCap data.} Multi-camera motion capture systems, such as Vicon~\citep{vicon}, 
track reflective markers to provide precise
hand motion annotations, \textit{including when hands are occluded in the
egocentric view}. We collected approximately 16 hours of hand-object
manipulation sequences from Noitom Robotics~\citep{noitom} and Bones~\citep{bones}, covering seated and standing
activities with varying levels of hand visibility. 
Motion capture is performed using micro-markers to maximize visibility of hand without using full-hand gloves.

\subsection{Data Annotation Pipeline}
\label{sec:soma-hand-annotation}

Our pipeline converts varied real-world annotations into a common
SOMA-Hand representation, accounting for differences in anatomical
joint definitions. We fit SOMA-Hand using calibrated video, supplied
3D hand landmarks, and SAM 3 segmentation masks. Anatomically
corresponding landmarks guide the fit, while 2D mask alignment
and 2D--3D consistency checks help assess agreement with the
observations.

SOMA-Hand's decoupled shape and pose parameters allow us to estimate
a single hand shape, including its proportions and overall size,
shared across each clip. With shape fixed, we optimize per-frame
poses, comprising finger articulation, wrist orientation, and wrist
position, with temporal constraints. This maintains consistent hand
geometry while capturing motion throughout the sequence. We first 
construct image-space fitting targets and estimate shared hand shape 
from selected frames, then fit poses across the full clip and refine 
difficult intervals before export.
% Figure~\ref{fig:soma-ann-pipeline} summarizes the workflow.

% To unify varied annotations into the SOMA-Hand format, our pipeline reconciles varying anatomical joint definitions. We resolve this by leveraging SOMA-Hand’s decoupled shape and pose parameterizations with robust 2D-3D consistency checks. This approach locks in a consistent per-clip hand identity (shape) while independently optimizing the per-frame motion (pose).

% We fit SOMA-Hand to real-world recordings using calibrated video,
% supplied 3D hand landmarks, and SAM 3 segmentation masks.
% The pipeline estimates a single \tc{size is part of shape?} hand shape and size for each clip,
% then \tc{are these just poses?} optimizes finger articulation, wrist orientation, and wrist
% position over time. This produces a consistent hand representation
% while allowing the pose to change throughout the sequence.

% \tc{need better terms than preparing, tell the goal}

% \textbf{Preparing the observations.}
% \tc{align is ambiguous}We align the supplied 3D landmarks with the video using camera
% calibration and rectify fisheye images when needed. SAM 3 provides
% masks of the visible hands, while the projected landmarks provide
% 2D keypoint targets. \tc{how does this work? Invalid is also ambiguous, maybe remove?} Invalid landmarks and unreliable mask regions
% are excluded from the corresponding alignment constraints.

\textbf{Constructing image-space fitting targets.}
We use camera calibration to transform the supplied 3D landmarks
into camera coordinates and project them onto the corresponding
video frames, producing 2D keypoint targets. Fisheye images are
rectified when needed, and landmark projection uses the intrinsics
of the rectified view. SAM 3 provides segmentation masks of the
visible hands. Forearm regions and ambiguous overlap between hands
are excluded from mask comparison so that they do not distort
the fitted hand silhouette.

% \textbf{Estimating a consistent hand shape.}
% We initialize hand size and wrist placement from the 3D landmarks.
% We then estimate hand shape from selected frames with clear
% visibility and diverse poses. The resulting shape and size are
% shared across the clip and held fixed during pose fitting,
% preventing frame-level noise from causing artificial changes
% in hand proportions. 
% \tc{first 3 sentences doesn't seem to connect.}
% We initialize hand size and wrist placement from the 3D landmarks.
% These landmarks also provide reference measurements
% of palm width and finger lengths.
% We then estimate hand shape from selected frames with clear
% visibility and diverse poses.
% During fitting, each selected frame has its own
% pose, while all frames share the same shape parameters.
% We alternate between adjusting these poses with shape fixed
% and adjusting the shared shape with poses fixed.
% Landmark alignment and agreement between the rendered hand
% silhouette and SAM 3 masks guide these updates, while dimension
% constraints discourage unrealistic changes in hand proportions.
% Combining evidence across frames reduces the dependency on any
% single pose or partially occluded observation.
% The resulting shape and size are shared across the clip and
% held fixed during pose fitting, preventing frame-level noise
% from causing artificial changes in hand proportions.

\textbf{Estimating a consistent hand shape.}
We estimate a single hand shape from selected frames with clear
visibility and diverse poses. The supplied 3D landmarks initialize
hand size and wrist placement, while measurements of palm width
and finger lengths constrain hand proportions during refinement.
Each selected frame has its own pose, but all frames share the
same shape parameters. We alternate between adjusting these poses
with shape fixed and adjusting the shared shape with poses fixed.
Landmark alignment and agreement between the rendered hand
silhouette and SAM 3 masks guide these updates, while the dimension
constraints discourage unrealistic changes in hand proportions.
Combining evidence across frames reduces dependence on any single
pose or partially occluded observation. The resulting shape,
including overall size, is then held fixed during pose fitting
across the full clip, preventing frame-level noise from causing
artificial changes in hand proportions.

% \hq{It'd be nice to provide a little more detail here, since it's a core part of the workflow.}

\textbf{Fitting hand motion.}
With shape fixed, we optimize each frame's pose using the supplied
3D landmarks, selected 2D keypoint correspondences, and SAM 3 masks.
The keypoints guide finger placement, while alignment between the
rendered hand silhouette and the masks constrains the visible
hand outline. We use anatomically corresponding keypoints and
exclude ambiguous matches, such as differing wrist definitions.
Temporal constraints connect neighboring frames to reduce jitter
while preserving alignment with the observations.

\textbf{Refinement and quality checks.}
Intervals with poor alignment or unstable motion receive additional
pose refinement, with shape remaining fixed. Updates are accepted
only when alignment and motion checks pass. Before export, we
validate frame coverage and model parameters. The final annotations
contain one shared hand shape and size per clip, together with
per-frame finger articulation and wrist placement. Their accuracy
remains dependent on the source observations.

% \begin{figure}
% \centering
% \includegraphics[width=\linewidth]{figures/soma_fit/pipeline.pdf}
% \caption{Annotation fitting for one hand sequence.
% The pipeline prepares calibrated video, landmarks, and SAM 3 masks;
% estimates a shared hand shape; fits per-frame poses; refines
% difficult intervals; and exports validated annotations.
% Blue denotes shape and size shared across the clip.
% Orange denotes finger articulation and wrist placement,
% which vary over time. Shape is held fixed during pose fitting.}
% \label{fig:soma-ann-pipeline}
% \end{figure}

%\subsection{Data Augmentation}

%Training uses crop-scale jitter, crop-center translation jitter, in-plane
%rotation, and brightness perturbation. Geometric augmentation applies
%consistent transformations to the image, hand geometry, camera information, and
%crop metadata. Each augmented crop retains the associated SOMA-Hand parameters
%and confidence masks.

% Drafted independently of sections/4_design.tex. Pipeline-specific labels
% allow both files to coexist until the manuscript structure is updated.
\section{End-to-End Inference Pipeline}
\label{sec:inference_pipeline}

% \hq{Too general, vague. Need a better name for this. Section covers everything, lacks a focus. If we move perspective cropping to data engine, perhpas we can be more focused on ViPE integration, global frame annotation production} -> trim this section significantly. 

\label{sec:pipeline_overview}

We provide an example modular inference pipeline that uses the \sysname model from Section~\ref{sec:method} to annotate egocentric video. Given a video and camera calibration, the pipeline detects hands, prepares image crops, reconstructs hand geometry, and assembles the predictions into adaptively smoothed sequences. A synchronized external camera trajectory enables conversion from camera-relative reconstructions to world-coordinate hand annotations for downstream policy learning.

\begin{figure}
\centering
\includegraphics[width=\linewidth]{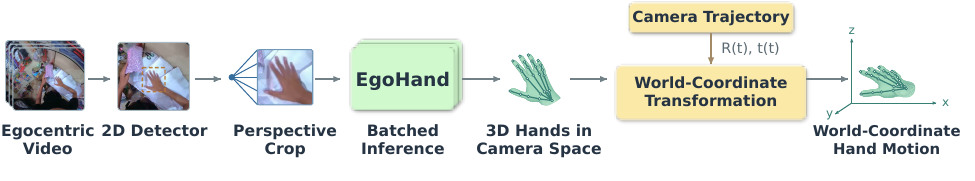}
\caption{Inference pipeline from egocentric video to world-coordinate hand motion, comprising hand detection, crop preparation, batched \sysname inference, and transformation of predictions to world coordinates. Fisheye regions are resampled into virtual pinhole views; pinhole images use standard crops. Predictions are assembled in frame order, their hand parameters are smoothed in an adaptive window. A synchronized external camera trajectory supplies the camera-to-world rotations and translations that place the hands in a shared world frame. Benchmark annotations may supply hand boxes in place of detections. The \sysname network is detailed in Section~\ref{sec:method}. Illustrations are schematic. 
% \hq{Perhaps this figure should come much earlier, as an overview section before describing egohand model}
% \hq{Could use an iteration. Remove top bold titles (caption already does that), remove gray legend at bottom (caption does it). reduce blank spaces, make the figure more compact} -> chaoyeh: figure is updated according to the suggestions :) 
}
\label{fig:egohand_inference_pipeline}
\end{figure}

Figure~\ref{fig:egohand_inference_pipeline} summarizes the workflow. Hand detection supplies bounding boxes for crop preparation. Pinhole images use standard crops, while fisheye images are resampled into virtual pinhole views with corresponding intrinsics. The pipeline batches these crops for \sysname inference and transforms predictions from virtual views back to the source-camera frame. Each prediction retains its source-frame identifier and handedness so that batch outputs can be assembled into left- and right-hand sequences. Adaptive postprocessing operates on the hand parameters before geometry is decoded for world-coordinate conversion and export.

The stages communicate through hand boxes, camera parameters, and reconstructed hand sequences. This organization allows the detector, camera-pose source, postprocessing, and execution infrastructure to be adapted while retaining the same reconstruction model.

\textbf{2D Hand Detection.} 
% \label{sec:pipeline_hand_detection} 
We provide an example 2D hand detector by fine-tuning
RT-DETRv2-S~\citep{lv2024rtdetrv2} on datasets listed in
Section~\ref{sec:dataset}, with bounding-box
annotations for both hands derived from the 3D annotations. 

% \textbf{\color{red}{Yanchen}}
% We select detections based on
% confidence and temporal overlap, convert them into padded square boxes,
% and apply temporal stabilization to reduce crop jitter.

\textbf{Fisheye Support.} 
% \label{sec:pipeline_perspective_cropping} 
To handle fisheye distortion, we apply \emph{perspective
cropping}~\citep{han2022umetrack} using a virtual pinhole camera that shares
the source camera's optical center and points toward the hand bounding-box
center. The resulting undistorted crop and virtual-camera intrinsics are 
then passed to \Sysname for reconstruction.

% \textbf{\color{red}{Xiufeng}}

\textbf{Adaptive Smoothing.}
% \label{sec:pipeline_postprocessing} 
Independent frame predictions can jitter, so
we apply offline smoothing to each hand's SOMA-Hand parameters using video
timestamps. Rotation and translation filtering adapts to local motion. 
We fit local quadratic motion curves, adjusting the smoothing 
window according to fitting error and estimated noise.
Windows narrow when a broader fit cannot capture local motion,
helping preserve motion changes while reducing jitter. Shape
and scale are also smoothed. We regenerate joints and meshes from the filtered
parameters.

\textbf{World-coordinate Reconstruction.} We use camera-to-world poses from
ViPE~\citep{huang2025vipe} or the recording to place hands in a shared world frame.
The trajectory must be aligned with the video frames and use the same camera
convention and metric scale as the hand predictions. For source-camera joint
$X^c_{t,j}$ and camera-to-world pose $(R^{c\to w}_t,t^{c\to w}_t)$,
\begin{equation}
X^w_{t,j} = R^{c\to w}_t X^c_{t,j} + t^{c\to w}_t.
\label{eq:pipeline_camera_to_world}
\end{equation}
An optional refinement step uses both hands jointly to adjust camera poses and
reduce shared world-space jitter, with additional smoothing of the resulting
keypoint trajectories. Without a camera trajectory, outputs remain camera-relative.

% TODO: Specify the released ViPE configuration for frame synchronization,
% coordinate conversion, and metric-scale alignment.

% \subsection{Large-Scale Inference}
\subsection{Hand Reconstruction at Production-scale}
\label{sec:pipeline_scaling}
% \textbf{\color{red}{Shao, Qingsi}}

We establish a hand reconstruction production pipeline that organizes video decoding, camera-pose estimation, hand detection, crop preparation, batched reconstruction, postprocessing, and LeRobot dataset assembly for annotation at scale. It targets collections exceeding 10,000 hours of egocentric video across tens of thousands of clips. At this scale, predictable resource use and efficient scheduling are essential to processing long or unusually large inputs reliably.

\textbf{Bounded memory footprint.} Decoding and processing operate on fixed-size chunks, so decoded-frame memory depends on chunk size rather than video length. Camera-pose estimation retains a fixed window of SLAM keyframes in GPU memory and retires older keyframes into a compact CPU-side trajectory record. Temporal hand postprocessing uses local windows rather than retaining full-track history on the GPU. Per-job resource caps and prompt cleanup bound resource use and allow multiple workers to share a machine.

\textbf{Batch processing and throughput.} Hand crops are pooled into fixed-size inference batches, and downstream postprocessing and GPU-to-CPU transfers are also batched to reduce per-item overhead. Forward passes use CUDA autocast with bfloat16, while numerically sensitive accumulation remains in float32. Each job processes multiple videos, amortizing startup, CUDA initialization, and model loading. In early production runs, batching videos per job reduced this fixed overhead by roughly an order of magnitude. Input downloads and output uploads run concurrently with computation, and results are streamed as each processing unit finishes.
% TODO(Shao, Qingsi): Provide the job configuration and timing basis for the
% fixed-overhead reduction; this is not an end-to-end model speedup claim.

\textbf{Deployment and output assembly.} Containers pin the CUDA toolchain and supported GPU architectures. Model checkpoints are included at build time, with a public default as fallback, to avoid repeated runtime downloads. Builds are identified by immutable SHA-256 digests, allowing outputs to be traced to the software version that produced them. NVIDIA OSMO workflows distribute inference to GPU pools and CPU-bound stages to CPU-rich pools. The final assembly stage associates hand annotations with their source frames and prepares LeRobot datasets for downstream policy-learning workflows.

\section{Evaluation}
\label{sec:eval}
% \textbf{\color{red}{Xiufeng}}

% We compare the reconstruction accuracy of \sysname and existing methods on the HOT3D~\citep{banerjee2025hot3d} and H2O~\citep{kwon2021h2o} datasets, and their computational cost on H2O.

\subsection{Experimental Setup}
\label{sec:eval_setup}

\textbf{Datasets.} We evaluate on the HOT3D~\citep{banerjee2025hot3d},
H2O~\citep{kwon2021h2o}, and ARCTIC~\citep{fan2023arctic} datasets.
None of these datasets is included in our training data. Our evaluation covers
approximately 2.11 hours of egocentric video from HOT3D, 1.06 hours from H2O,
and 0.23 hours from ARCTIC, all at 30\,fps.

\textbf{Baselines.} On H2O, we compare with single-image methods
HaMeR~\citep{pavlakos2024hamer} and WiLoR~\citep{potamias2025wilor} and video-based
methods HaWoR~\citep{zhang2025hawor} and Dyn-HaMR~\citep{yu2025dynhamr}. None of
these methods was trained on H2O. On HOT3D and ARCTIC, we compare with HaMeR and Dyn-HaMR,
excluding WiLoR and HaWoR because their training data include HOT3D and ARCTIC. 
% On ARCTIC, we compare with HaMeR and Dyn-HaMR; WiLoR and HaWoR are excluded because their training data include ARCTIC. 
In our evaluation, we provide HaWoR and Dyn-HaMR with ground-truth camera poses
to exclude camera-pose estimation error from the comparison. Our evaluation
compares hand reconstruction from ground-truth hand boxes and does not assess
completion of missing hand trajectories. We therefore omit HaWoR's separate
motion-infilling module. We exclude Dyn-HaMR's hand motion prior
(HMP) refinement because we could not run this stage successfully with the
official code.

\textbf{Evaluation protocol.} All methods receive ground-truth hand boxes to
evaluate reconstruction independently of detection. \sysname uses FP16 inference
with 256 \texttimes{} 256 inputs. On H2O, all methods use hand crops from the RGB images.
On HOT3D, \sysname uses perspective crops from fisheye images.
On ARCTIC, all methods use hand crops
from globally undistorted RGB images, with boxes obtained by projecting
ground-truth MANO mesh vertices. We apply adaptive smoothing to \sysname predictions
unless a result is labeled ``without smoothing.''
To compare against MANO annotations, we map
\sysname's predicted SOMA-Hand surfaces to MANO topology, fit a MANO mesh
to the mapped surface, and extract 21 evaluation landmarks. 
% \hq{Shall we briefly explain this: a reverse process of section~\ref{sec:soma-hand-annotation}}

% TODO(Xiufeng): Add training hours separately for synthetic data, XDOF, Mecka Stereo, Noitom, and Bones when supplied.

\textbf{Metrics.}
\label{sec:eval_metrics}
We report the mean and p90 of wrist-relative MPJPE, PA-MPJPE, and root translation
error (RTE) in mm, 2D EPE in pixels, and acceleration error in mm/frame\textsuperscript{2}. See Appendix~\ref{sec:eval_metric_details} for detailed definition.

\subsection{Evaluation Results}

%Tables~\ref{tab:egohand_h2o_results}--\ref{tab:arctic_results} compare mean and p90 errors on H2O, HOT3D, and ARCTIC across MPJPE, PA-MPJPE, RTE, 2D EPE, and acceleration error.

Tables~\ref{tab:egohand_h2o_results}--\ref{tab:arctic_results} compare \sysname at 60k training steps with baseline methods, reporting mean and p90 errors on H2O, HOT3D, and ARCTIC across MPJPE, PA-MPJPE, RTE, 2D EPE, and acceleration error.

\Needspace{24\baselineskip}
\textbf{H2O.} \sysname achieves the lowest mean and p90 MPJPE, RTE, and acceleration error. WiLoR retains the lowest mean and p90 PA-MPJPE and 2D EPE. Its two-stage reconstruction refines an initial hand estimate using image-aligned features~\citep{potamias2025wilor}, at a higher inference cost (Section~\ref{sec:runtime}, Table~\ref{tab:runtime}).

% \hq{These tables 1-3 are difficult. Convention is to transpose it. Each column is a metric, each row is a method. Bottom row is ours (best method)}
\begin{table}[!htbp]
\centering\small
\setlength{\tabcolsep}{2.3pt}
\caption{Reconstruction accuracy on H2O. Lower is better for all metrics.}
\label{tab:egohand_h2o_results}
% Baseline values retained from figures/plot_data.csv; both EgoHand variants from adaptive_smoothing_results_provenance.json.
\begin{tabularx}{\linewidth}{@{}l|CC|CC|CC|CC|CC@{}}
% \toprule
& \multicolumn{2}{c|}{MPJPE $\downarrow$}
& \multicolumn{2}{c|}{PA-MPJPE $\downarrow$}
& \multicolumn{2}{c|}{RTE $\downarrow$}
& \multicolumn{2}{c|}{2D EPE $\downarrow$}
& \multicolumn{2}{c}{Accel. error $\downarrow$} \\
& \multicolumn{2}{c|}{(mm)}
& \multicolumn{2}{c|}{(mm)}
& \multicolumn{2}{c|}{(mm)}
& \multicolumn{2}{c|}{(px)}
& \multicolumn{2}{c}{(mm/frame\textsuperscript{2})} \\
% \cmidrule(lr){2-3}\cmidrule(lr){4-5}\cmidrule(lr){6-7}\cmidrule(lr){8-9}\cmidrule(lr){10-11}
Method & Mean & p90 & Mean & p90 & Mean & p90 & Mean & p90 & Mean & p90 \\
\midrule
HaMeR~\citep{pavlakos2024hamer}
& 19.35 & 28.04 & 7.13 & 9.61 & 37.25 & 71.41 & 11.15 & 15.83 & 4.53 & 7.58 \\
HaWoR~\citep{zhang2025hawor}
& 17.78 & 24.05 & 7.18 & 9.75 & 35.63 & 55.71 & 17.08 & 22.08 & 2.82 & 5.52 \\
WiLoR~\citep{potamias2025wilor}
& 17.77 & 25.32
& \textcolor{green!50!black}{\textbf{6.06}}
& \textcolor{green!50!black}{\textbf{8.16}}
& 30.70 & 56.78
& \textcolor{green!50!black}{\textbf{8.96}}
& \textcolor{green!50!black}{\textbf{12.37}}
& 3.87 & 6.50 \\
Dyn-HaMR~\citep{yu2025dynhamr}
& 19.32 & 27.97 & 7.14 & 9.61 & 37.18 & 71.55 & 15.50 & 22.90 & 4.36 & 7.28 \\
\midrule
{\sysname\ (Ours) w/o smoothing}
& 13.24 & 18.91 & 6.91 & 8.97 & 25.70 & 42.26 & 9.61 & 12.68 & 3.92 & 6.26 \\
\sysname\ (Ours) w/ smoothing
& \textcolor{green!50!black}{\textbf{13.12}}
& \textcolor{green!50!black}{\textbf{18.75}}
& 6.87 & 8.92
& \textcolor{green!50!black}{\textbf{25.63}}
& \textcolor{green!50!black}{\textbf{42.10}}
& 9.59 & 12.66
& \textcolor{green!50!black}{\textbf{1.90}}
& \textcolor{green!50!black}{\textbf{3.92}} \\
% \bottomrule
\end{tabularx}
\end{table}

% \Needspace{16\baselineskip}
\textbf{HOT3D.} \sysname achieves lower mean and p90 errors than HaMeR and Dyn-HaMR across all five metrics.

\begin{table}[!htbp]
\centering\small
\setlength{\tabcolsep}{2.3pt}
\caption{Reconstruction accuracy on HOT3D. Lower is better for all metrics.}
\label{tab:egohand_hot3d_results}
% Baseline values retained from figures/plot_data.csv; both EgoHand variants from adaptive_smoothing_results_provenance.json.
\begin{tabularx}{\linewidth}{@{}l|CC|CC|CC|CC|CC@{}}
% \toprule
& \multicolumn{2}{c|}{MPJPE $\downarrow$}
& \multicolumn{2}{c|}{PA-MPJPE $\downarrow$}
& \multicolumn{2}{c|}{RTE $\downarrow$}
& \multicolumn{2}{c|}{2D EPE $\downarrow$}
& \multicolumn{2}{c}{Accel. error $\downarrow$} \\
& \multicolumn{2}{c|}{(mm)}
& \multicolumn{2}{c|}{(mm)}
& \multicolumn{2}{c|}{(mm)}
& \multicolumn{2}{c|}{(px)}
& \multicolumn{2}{c}{(mm/frame\textsuperscript{2})} \\
% \cmidrule(lr){2-3}\cmidrule(lr){4-5}\cmidrule(lr){6-7}\cmidrule(lr){8-9}\cmidrule(lr){10-11}
Method & Mean & p90 & Mean & p90 & Mean & p90 & Mean & p90 & Mean & p90 \\
\midrule
HaMeR~\citep{pavlakos2024hamer}
& 58.00 & 79.47 & 9.47 & 13.02 & 51.92 & 89.60 & 22.88 & 36.41 & 5.92 & 10.04 \\
Dyn-HaMR~\citep{yu2025dynhamr}
& 58.16 & 79.78 & 9.64 & 13.25 & 53.51 & 94.91 & 22.95 & 36.50 & 5.83 & 9.88 \\
\midrule
{\sysname\ (Ours) w/o smoothing}
& 18.93 & 26.00 & 9.10 & 11.41 & 36.20 & 57.87 & 16.43
& \textcolor{green!50!black}{\textbf{22.06}}
& 4.39 & 7.29 \\
{\sysname\ (Ours) w/ smoothing}
& \textcolor{green!50!black}{\textbf{18.82}}
& \textcolor{green!50!black}{\textbf{25.78}}
& \textcolor{green!50!black}{\textbf{9.04}}
& \textcolor{green!50!black}{\textbf{11.32}}
& \textcolor{green!50!black}{\textbf{36.13}}
& \textcolor{green!50!black}{\textbf{57.63}}
& \textcolor{green!50!black}{\textbf{16.42}}
& 22.07
& \textcolor{green!50!black}{\textbf{1.63}}
& \textcolor{green!50!black}{\textbf{3.10}} \\
% \bottomrule
\end{tabularx}
\end{table}

% \Needspace{24\baselineskip}
\textbf{ARCTIC.} \sysname achieves lower mean and p90 errors than HaMeR and Dyn-HaMR
across all five metrics, with and without adaptive smoothing. Smoothing reduces
mean acceleration error from 7.64 to 2.64\,mm/frame\textsuperscript{2} while also
improving the spatial metrics. With smoothing, \sysname reduces mean MPJPE, RTE, and acceleration error relative to HaMeR by 34.9\%,
51.4\%, and 67.5\%, respectively.
Although \sysname processes 256 \texttimes{} 256 hand crops, ARCTIC's 2D EPE
is measured in the full resolution 2800 \texttimes{} 2000 image coordinates
after undistortion. A given relative localization error therefore spans more
pixels than in the lower-resolution H2O and HOT3D images, contributing to
ARCTIC's larger 2D~EPE.

\begin{table}[!bhtp]
\centering\small
\setlength{\tabcolsep}{2.3pt}
\caption{Reconstruction accuracy on ARCTIC. Lower is better for all metrics.}
\label{tab:arctic_results}
% Metric values from arctic_results_provenance.json.
\begin{tabularx}{\linewidth}{@{}l|CC|CC|CC|CC|CC@{}}
% \toprule
& \multicolumn{2}{c|}{MPJPE $\downarrow$}
& \multicolumn{2}{c|}{PA-MPJPE $\downarrow$}
& \multicolumn{2}{c|}{RTE $\downarrow$}
& \multicolumn{2}{c|}{2D EPE $\downarrow$}
& \multicolumn{2}{c}{Accel. error $\downarrow$}\\
& \multicolumn{2}{c|}{(mm)}
& \multicolumn{2}{c|}{(mm)}
& \multicolumn{2}{c|}{(mm)}
& \multicolumn{2}{c|}{(px)}
& \multicolumn{2}{c}{(mm/frame\textsuperscript{2})} \\
% \cmidrule(lr){2-3}\cmidrule(lr){4-5}\cmidrule(lr){6-7}\cmidrule(lr){8-9}\cmidrule(lr){10-11}
Method & Mean & p90 & Mean & p90 & Mean & p90 & Mean & p90 & Mean & p90 \\
\midrule
HaMeR~\citep{pavlakos2024hamer}
& 26.44 & 38.05 & 10.33 & 15.18 & 48.35 & 87.70 & 69.59 & 100.68 & 8.12 & 13.30 \\
Dyn-HaMR~\citep{yu2025dynhamr}
& 26.60 & 38.41 & 10.46 & 15.40 & 50.40 & 91.61 & 69.34 & 100.94 & 7.72 & 12.93 \\
\midrule
{\sysname\ (Ours) w/o smoothing}
& 17.53 & 24.78 & 9.04 & 12.27 & 23.88 & 42.26 & 45.04 & 66.82 & 7.64 & 12.88 \\
{\sysname\ (Ours) w/ smoothing}
& \textcolor{green!50!black}{\textbf{17.21}}
& \textcolor{green!50!black}{\textbf{24.42}}
& \textcolor{green!50!black}{\textbf{8.95}}
& \textcolor{green!50!black}{\textbf{12.13}}
& \textcolor{green!50!black}{\textbf{23.51}}
& \textcolor{green!50!black}{\textbf{41.49}}
& \textcolor{green!50!black}{\textbf{44.84}}
& \textcolor{green!50!black}{\textbf{66.11}}
& \textcolor{green!50!black}{\textbf{2.64}}
& \textcolor{green!50!black}{\textbf{4.74}} \\
% \bottomrule
\end{tabularx}
\end{table}

\subsection{Ablation Study}
\label{sec:ablation}

We examine the benefits of ray positional encoding (RayPE), bounding-box
conditioning (CLIFF), and the 2D joint prediction branch. Table~\ref{tab:component_ablation} reports results at 40k training steps without adaptive smoothing. %We emphasize MPJPE, RTE, and acceleration error, and use PA-MPJPE and 2D EPE to further assess hand geometry and image alignment.

\begin{table}[htbp]
\centering\small
\setlength{\tabcolsep}{2.3pt}
\caption{Component ablations on H2O and HOT3D without adaptive smoothing. All entries are mean errors.
%RayPE denotes ray positional encoding, CLIFF denotes bounding-box conditioning, and 2D denotes the 2D joint prediction branch.
}
\label{tab:component_ablation}
\footnotesize
\setlength{\tabcolsep}{1.5pt}
\renewcommand{\arraystretch}{1.15}
\begin{tabularx}{\linewidth}{@{}ccc|CCCCC|CCCCC@{}}
& & & \multicolumn{5}{c|}{\textbf{H2O}} & \multicolumn{5}{c}{\textbf{HOT3D}} \\
\cmidrule(lr){4-8}\cmidrule(lr){9-13}
RayPE & CLIFF & 2D & MPJPE & \makecell{PA-\\MPJPE} & RTE & \makecell{2D\\EPE} & \makecell{Accel.\\error} & MPJPE & \makecell{PA-\\MPJPE} & RTE & \makecell{2D\\EPE} & \makecell{Accel.\\error} \\
 &  &  & (mm) $\downarrow$ & (mm) $\downarrow$ & (mm) $\downarrow$ & (px) $\downarrow$ & $\frac{\mathrm{mm}}{\mathrm{frame}^{2}}\!\downarrow$ & (mm) $\downarrow$ & (mm) $\downarrow$ & (mm) $\downarrow$ & (px) $\downarrow$ & $\frac{\mathrm{mm}}{\mathrm{frame}^{2}}\!\downarrow$ \\
\midrule
\xmark & \xmark & \xmark & 18.45 & 8.20 & 27.03 & 13.40 & 4.56 & 26.15 & 9.10 & 38.30 & 26.63 & 6.04 \\
\cmark & \xmark & \xmark & 15.74 & 8.31 & \textcolor{green!50!black}{\textbf{25.09}} & 12.90 & 3.91 & \textcolor{green!50!black}{\textbf{18.17}} & 8.99 & \textcolor{green!50!black}{\textbf{35.19}} & 17.33 & 4.91 \\
\xmark & \cmark & \xmark & 16.06 & 8.38 & 26.16 & 12.55 & 3.90 & 22.09 & 9.29 & 36.03 & 21.51 & 5.30 \\
\xmark & \xmark & \cmark & 18.89 & 7.56 & 25.12 & 10.53 & 5.67 & 23.45 & 9.17 & 41.51 & 18.78 & 6.67 \\
\cmark & \cmark & \xmark & 15.85 & 8.24 & 25.83 & 12.67 & \textcolor{green!50!black}{\textbf{3.54}} & 18.41 & 9.00 & 35.63 & 16.91 & \textcolor{green!50!black}{\textbf{4.43}} \\
\cmark & \xmark & \cmark & 14.55 & 7.23 & 25.77 & 9.65 & 4.28 & 19.05 & \textcolor{green!50!black}{\textbf{8.94}} & 39.14 & 16.61 & 4.68 \\
\xmark & \cmark & \cmark & 14.30 & 7.07 & 25.62 & 9.67 & 4.30 & 19.18 & 9.19 & 38.72 & 17.09 & 5.14 \\
\midrule
\cmark & \cmark & \cmark & \textcolor{green!50!black}{\textbf{13.31}} & \textcolor{green!50!black}{\textbf{6.97}} & 26.00 & \textcolor{green!50!black}{\textbf{9.51}} & 4.06 & 18.90 & 9.02 & 38.44 & \textcolor{green!50!black}{\textbf{16.47}} & 4.69 \\
\end{tabularx}
\end{table}

\textbf{Ray positional encoding.}
With CLIFF and the 2D branch disabled, adding
RayPE reduces MPJPE from 18.45 to 15.74\,mm on H2O and from 26.15 to
18.17\,mm on HOT3D, while also lowering RTE and acceleration error on
both datasets. These comparisons demonstrate its contribution
to pose accuracy across different component configurations.

\textbf{Bounding-box conditioning.}
With RayPE and the 2D branch disabled, adding CLIFF reduces MPJPE from 18.45 to 16.06\,mm on H2O and from 26.15 to 22.09\,mm on HOT3D, with improvements in RTE and acceleration error.
With RayPE and the 2D branch enabled, adding CLIFF further reduces
MPJPE from 14.55 to 13.31\,mm on H2O and from 19.05 to 18.90\,mm
on HOT3D. These results show that bounding-box conditioning complements
ray encoding and 2D joint prediction to improve hand-pose accuracy.

\textbf{2D joint prediction.}
With RayPE and CLIFF enabled, adding the 2D branch reduces MPJPE
from 15.85 to 13.31\,mm, PA-MPJPE from 8.24 to 6.97\,mm, and
2D EPE from 12.67 to 9.51\,px on H2O. On HOT3D, 2D EPE also
decreases from 16.91 to 16.47\,px. These gains come with higher RTE and
acceleration error on both datasets, and a small increase in HOT3D
MPJPE from 18.41 to 18.90\,mm. Since correspondence-based refinement
adjusts lateral translation while retaining the predicted depth,
better image alignment does not necessarily imply lower RTE. These results highlight complementary benefits in camera-aware pose
estimation and image alignment. %We combine the
%components to balance hand reconstruction accuracy and image alignment
%across datasets. 

%The complete model achieves the lowest H2O MPJPE among the evaluated configurations, although selected ablations yield lower translation or acceleration errors, and lower MPJPE on HOT3D.

%\Needspace{20\baselineskip}
\subsection{Inference Latency}
\label{sec:runtime}
Table~\ref{tab:runtime} reports per-frame latency on an NVIDIA L40 GPU, with an
additional \sysname measurement on an NVIDIA RTX PRO 6000 Blackwell GPU.
For each method and GPU configuration, we load the model once and run the pipeline
repeatedly in the same Python process. We exclude initial warm-up runs and report
mean latency per source frame.
For \sysname, inference includes batch preparation (batch size 32), the forward pass,
2D keypoint projection, and mirroring right-hand predictions back from the
model's left-hand convention. It excludes 2D hand detection, cropping, and rendering.
Post-processing comprises adaptive smoothing for \sysname.
Dyn-HaMR~\citep{yu2025dynhamr} initializes hand predictions with HaMeR; its
post-processing consists of offline temporal optimization.

\begin{table}[htbp]
\centering\small
\caption{Per-frame latency on the indicated NVIDIA GPUs.}
\label{tab:runtime}
\begin{tabular}{ll|cc|c}
Method & GPU & \makecell{Inference $\downarrow$\\(ms/frame)} & \makecell{Post-processing $\downarrow$\\(ms/frame)} & \makecell{Total $\downarrow$\\(ms/frame)} \\
\midrule
HaMeR~\citep{pavlakos2024hamer} & L40 & 26.00 & --- & 26.00 \\
WiLoR~\citep{potamias2025wilor} & L40 & 27.43 & --- & 27.43 \\
HaWoR~\citep{zhang2025hawor} & L40 & 26.64 & --- & 26.64 \\
Dyn-HaMR~\citep{yu2025dynhamr} & L40 & 26.00 & 144.03 & 170.03 \\
\midrule
\sysname\ (ours) & L40 & \textcolor{green!50!black}{\textbf{18.35}} & \textcolor{green!50!black}{\textbf{2.90}} & \textcolor{green!50!black}{\textbf{21.25}} \\
\midrule
\sysname\ (ours) & \makecell[l]{RTX PRO 6000} & \textcolor{green!50!black}{\textbf{9.01}} & \textcolor{green!50!black}{\textbf{1.86}} & \textcolor{green!50!black}{\textbf{10.87}} \\
\end{tabular}
\end{table}

% \subsection{Ablation Study}
% \label{sec:ablation}

% Table~\ref{tab:component_ablation} compares earlier model configurations trained for 40,000 steps on the same synthetic and external vendor data sources: a baseline and variants with ray positional encoding (RayPE) or the separate 2D prediction branch. The 2D branch is paired with logarithmic-depth prediction. Other training and preprocessing settings are not fully matched across rows, so these comparisons do not isolate individual component contributions.

% \begin{table}[H]
% \centering\small
% \caption{Model-variant comparisons after 40,000 training steps.}
% \label{tab:component_ablation}
% \begin{tabular}{lrrrr}
% \toprule
% & \multicolumn{2}{c}{H2O} & \multicolumn{2}{c}{HOT3D} \\
% \cmidrule(lr){2-3}\cmidrule(lr){4-5}
% Configuration & MPJPE (mm) $\downarrow$ & RTE (mm) $\downarrow$ & MPJPE (mm) $\downarrow$ & RTE (mm) $\downarrow$ \\
% \midrule
% Baseline & 17.61 & 39.35 & 20.26 & 43.31 \\
% + RayPE & 17.21 & \textbf{35.44} & 26.04 & 47.04 \\
% + 2D branch and log depth & \textbf{16.48} & 37.50 & \textbf{19.59} & \textbf{39.65} \\
% \bottomrule
% \end{tabular}
% \end{table}

% The combined 2D-branch and log-depth variant has lower MPJPE and RTE than the baseline on both datasets; H2O 2D EPE also decreases from 14.11 to 10.26 pixels. RayPE has lower MPJPE and RTE than the baseline on H2O but higher errors on HOT3D in these runs.

\subsection{Human Preference Study}
\label{sec:qualitative}

We conducted a blinded A/B study comparing the visual accuracy of 2D overlays of \sysname predictions and dataset-provided hand poses. We randomly selected 200 manipulation clips from each of the three public egocentric datasets: Apple EgoDex~\citep{hoque2026egodex}, Ropedia Xperience-10M~\citep{ropedia_xperience_10m}, and Lightwheel EgoStandard, part of EgoSuite-Open100K~\citep{lightwheel_egostandard}. The 2D hand bounding boxes provided to \sysname were derived by projecting the datasets' 3D hand poses and applying a 30\% padding. 

We overlaid the 2D projections of the reconstructed 3D joints on the videos, presented side-by-side with method identities concealed. We used three independently randomized assignment sets. Each of the 600 clip pairs was independently rated by three graders with hand-annotation experience, yielding 1,800 primary ratings. Graders scored preference on a five-point scale and rated each side from 1 (unusable) to 5 (excellent). A majority required at least two graders to agree on direction, grouping slight and clear preferences; a three-way split was considered inconclusive.

% \input{fig_latex/user_study}
% \begin{table}[htbp]
% \centering\small
% \setlength{\tabcolsep}{6pt}
% \caption{Majority preferences and mean quality ratings across 200 clips per dataset. Bold green values indicate the largest preference percentage and higher mean quality within each dataset.}
% \label{tab:qualitative_preferences}
% \label{tab:qualitative_scores}
% \begin{tabular}{l|ccc}
% & EgoDex & Xperience-10M & EgoStandard \\
% & \citep{hoque2026egodex}
% & \citep{ropedia_xperience_10m}
% & \citep{lightwheel_egostandard} \\
% \midrule
% \multicolumn{4}{l}{\textit{Majority preference (\%)}} \\
% \sysname\
% & \textcolor{green!50!black}{\textbf{77.0}}
% & \textcolor{green!50!black}{\textbf{95.5}}
% & 4.0 \\
% No meaningful difference
% & 14.0 & 2.0
% & \textcolor{green!50!black}{\textbf{91.0}} \\
% Dataset poses
% & 2.0 & 1.0 & 2.5 \\
% No majority
% & 7.0 & 1.5 & 2.5 \\
% \midrule
% \multicolumn{4}{l}{\textit{Mean quality (1--5) $\uparrow$}} \\
% \sysname
% & \textcolor{green!50!black}{\textbf{4.26}}
% & \textcolor{green!50!black}{\textbf{3.54}}
% & \textcolor{green!50!black}{\textbf{4.70}} \\
% Dataset poses
% & 3.54 & 2.50 & 4.66 \\
% \end{tabular}
% \end{table}

\begin{table}[htbp]
\centering
\begin{minipage}[t]{0.49\textwidth}
\vspace{0pt}
\centering\small
\setlength{\tabcolsep}{2pt}
\caption{Majority preference (\%, $\uparrow$).}
\label{tab:user_study_preference}
\begin{tabular}{l|c|c|c}
& EgoDex & Xperience-10M & EgoStandard \\
\midrule
Dataset poses
& 2.0 & 1.0 & 2.5 \\
Tie / Inconclusive
& 14.0 / 7.0 & 2.0 / 1.5
& \textcolor{green!50!black}{\textbf{91.0}} / 2.5 \\
% Tie
% & 14.0 & 2.0
% & \textcolor{green!50!black}{\textbf{91.0}} \\
% No majority
% & 7.0 & 1.5 & 2.5 \\
\sysname (Ours)
& \textcolor{green!50!black}{\textbf{77.0}}
& \textcolor{green!50!black}{\textbf{95.5}}
& 4.0 \\
\end{tabular}
\end{minipage}
\hfill
\begin{minipage}[t]{0.49\textwidth}
\vspace{0pt}
\centering\small
\setlength{\tabcolsep}{2pt}
\caption{Mean user rated quality (1--5, $\uparrow$).}
\label{tab:user_study_quality}
\begin{tabular}{l|c|c|c}
& EgoDex & Xperience-10M & EgoStandard \\
\midrule
Dataset poses
& 3.54 & 2.50 & 4.66 \\
\sysname (Ours)
& \textcolor{green!50!black}{\textbf{4.26}}
& \textcolor{green!50!black}{\textbf{3.54}}
& \textcolor{green!50!black}{\textbf{4.70}} \\
\end{tabular}
\end{minipage}
\end{table}

\Needspace{7\baselineskip}
\sysname achieved clear gains over the supplied annotations, earning majority preference on 77.0\% of EgoDex clips and 95.5\% of Xperience-10M clips (Table~\ref{tab:user_study_preference}). On EgoStandard, which provides high-quality hand annotations, \sysname achieved comparable visual quality, with a majority of graders reporting a tie or an inconclusive preference on 93.5\% of clips. Mean quality ratings followed the same pattern (Table~\ref{tab:user_study_quality}). These results demonstrate \sysname's effectiveness for producing high-quality hand annotations across diverse egocentric recordings.

\section{Conclusion}
\label{sec:discussion}
\label{sec:conclusion}

We presented \sysname, a monocular 3D hand reconstruction system for scalable
annotation of egocentric video. Trained on 14 million images spanning synthetic
scenes, in-the-wild recordings, and motion capture unified through SOMA-Hand,
the model combines camera geometry cues with auxiliary 2D joint prediction
and translation refinement. Zero-shot evaluations on HOT3D, H2O, and ARCTIC
demonstrate strong reconstruction accuracy, with adaptive smoothing further
reducing acceleration error. Component ablations characterize the benefits and
tradeoffs of camera conditioning and 2D prediction. A blinded human study shows
a preference for \sysname's projected poses over supplied EgoDex and
Xperience-10M annotations, and predominantly no meaningful difference on
EgoStandard. Together with the modular video-processing pipeline, these results
support large-scale egocentric hand annotation.
Nevertheless, several fundamental and practical challenges remain:

\textbf{Monocular Ambiguity \& Occlusion.} Monocular inputs inherently suffer
from depth-scale ambiguity. In single-frame egocentric settings, severe self-
and object occlusions require predicting occluded joints primarily from
learned priors. To mitigate depth ambiguity and enhance robustness under
occlusion, we are investigating stereo setups and wrist-mounted cameras to
introduce additional geometric constraints and viewpoint diversity.

\textbf{Upstream Sensitivity.} Detection errors in upstream bounding boxes
propagate through crop-based geometric features, and inaccurate camera
intrinsics impair 3D reconstruction. To reduce upstream dependencies, we are
exploring end-to-end detection-reconstruction frameworks and leveraging stereo
cues to refine 2D hand localization.

Future work will also explore the impact of annotation quality on downstream
robot learning.

\clearpage

\clearpage
\setcitestyle{numbers}
\bibliographystyle{plainnat}
\bibliography{references}

\clearpage
\appendix
\section{Contributors and Acknowledgments}
\label{sec:contributors}

% Inspired from Deepmind's Gemini: https://arxiv.org/abs/2312.11805

% Start a local group to limit changes
{
\footnotesize
\setlength{\columnsep}{12pt}
\setlength{\parindent}{0pt}
\linespread{0.96}\selectfont

\begin{multicols}{3}
\subsection*{\normalsize{Core Contributors}}

% These are individuals who have made significant impact to Isaac Lab Framework. They are engaged with Isaac Lab (not (just) as a user but as a core contributor) and handle day-to-day working of the framework. Their contributions go beyond code, i.e. they are heavily involved in the project's mission, PR reviewing, issue handling, and engaging with the community. 

Lin Duan, \textit{NVIDIA} \\
Jiefeng Li, \textit{NVIDIA} \\
Xiufeng Xie, \textit{NVIDIA} \\
Yu-Hsiang Huang, \textit{NVIDIA} \\
Yanchen Liu, \textit{NVIDIA} \\
Shao Su, \textit{NVIDIA} \\
Hang Qiu, \textit{NVIDIA} \\
Yihan Pang, \textit{NVIDIA, UIUC} \\
Tianyi Hu, \textit{NVIDIA, Duke University} \\

\columnbreak
\subsection*{\normalsize{Contributors}}
Edy Lim, \textit{NVIDIA} \\
Hui Kang, \text{NVIDIA} \\
Juana Du, \text{NVIDIA} \\
Jun Saito, \textit{NVIDIA} \\
Megha Maheshwari, \textit{NVIDIA} \\

\subsection*{\normalsize{Leadership}}
David Chu, \textit{NVIDIA} \\
Karen Zhu, \textit{NVIDIA} \\
Spencer Huang, \textit{NVIDIA} \\
Trisha Tripathi, \textit{NVIDIA} \\

\columnbreak
\subsection*{\normalsize{Core Leadership}}
Chao-Yeh Chen, \textit{NVIDIA} \\
Tiffany Chen, \textit{NVIDIA} \\
Umar Iqbal, \textit{NVIDIA} \\
Qingsi Wang, \textit{NVIDIA} \\

\end{multicols}
}
\vspace{-10pt}
\noindent\hrulefill

The roles are defined as follows:

\textbf{Core Contributor:} Individuals who had a very significant and sustained impact throughout the project, across multiple subsystems, with accountability for key design and implementation decisions. Core contributors are listed in order of merit.

\textbf{Contributor:} Individuals who made important contributions to the project. Contributors are listed in alphabetical order. 

\textbf{Leadership:} Individuals who contributed to the organizational and technical direction of efforts related to \sysname{}. Leadership members are listed in alphabetical order.

\textbf{Core Leadership:} Individuals with primary responsibility for the organizational and technical direction of \sysname{}. Core leadership members are listed in alphabetical order.

\noindent\hrulefill

\subsection*{Acknowledgments}

We are especially grateful to Ben Levin, Ryan Punamiya, Aly Khater, Danfei Xu, Danyi Chen, Anna Vargas De Inoue, Yuke Zhu, Cyrus Hogg, Sandeep Desai, Ariana Lerena Riofrio, Naema Bhatti, Yan Chang, Xinghao Zhu, and Soha Pouya for their constructive feedback and valuable guidance.

We also acknowledge the use of large language models, including OpenAI GPT-6, to improve the manuscript’s language and clarity. All technical content and analyses were developed by the authors. Recognizing that these models may introduce biases, errors, or omissions, the authors take full responsibility for the manuscript’s accuracy and appropriateness.

\section{Metric Definitions}
\label{sec:eval_metric_details}

On HOT3D, H2O, and ARCTIC, we evaluate the reconstructed hand's joint configuration, wrist location, and motion using \textbf{mean per-joint position error (MPJPE)}, \textbf{root translation error (RTE)}, and \textbf{acceleration error (Accel.)}, respectively. We also report \textbf{Procrustes-aligned MPJPE (PA-MPJPE)} and \textbf{2D endpoint error (2D EPE)} as secondary metrics. Lower values indicate better performance for all five metrics.

We represent each evaluated hand using $N=21$ landmarks, including the wrist. For one hand in one frame, let $P_j,G_j\in\mathbb{R}^3$ denote predicted and ground-truth joint positions in the source-camera frame, expressed in millimeters, with wrist index $j=0$. Define wrist-relative coordinates $p_j=P_j-P_0$ and $g_j=G_j-G_0$.

\textbf{MPJPE} measures the mean distance between corresponding wrist-relative joints:
\begin{equation}
E_{\mathrm{MPJPE}}=\frac{1}{N}\sum_{j=0}^{N-1}
\left\|p_j-g_j\right\|_2.
\end{equation}
Subtracting each hand's own wrist removes global translation. No rotation or scale alignment is applied, so this metric retains errors in hand orientation, scale, and articulation. The wrist contributes zero error but remains in the 21-joint average. MPJPE is reported in millimeters.

\textbf{RTE} measures the distance between the predicted and ground-truth wrists in camera coordinates:
\begin{equation}
E_{\mathrm{RTE}}=\left\|P_0-G_0\right\|_2.
\end{equation}
It is reported in millimeters and measures absolute hand localization using the wrist of the evaluated joint set.

\textbf{Acceleration error} measures agreement with ground-truth hand motion over time. For each hand and clip, we transform predicted and ground-truth joints into a common world frame using the dataset's camera-to-world transforms, then subtract each sequence's wrist independently at every frame. Writing the transformed positions in millimeters as $P^w_{t,j}$ and $G^w_{t,j}$, define wrist-relative world-space joints $\bar P_{t,j}=P^w_{t,j}-P^w_{t,0}$ and $\bar G_{t,j}=G^w_{t,j}-G^w_{t,0}$. With the centered second difference $\Delta^2\bar P_{t,j}=\bar P_{t+1,j}-2\bar P_{t,j}+\bar P_{t-1,j}$, and analogously for ground truth, the error at the middle frame is
\begin{equation}
E_{\mathrm{Accel},t}=\frac{1}{N}\sum_{j=0}^{N-1}
\left\|\Delta^2\bar P_{t,j}-\Delta^2\bar G_{t,j}\right\|_2.
\end{equation}
We report this quantity in \textbf{mm/frame\textsuperscript{2}}, using consecutive steps in the ordered evaluation sequence without dividing by elapsed time in seconds. Comparisons therefore require consistent frame rates and sampling intervals. The metric measures wrist-relative motion agreement, excluding acceleration of the global wrist trajectory. A smoother prediction alone need not have a lower error.

\textbf{PA-MPJPE} measures joint error after aligning prediction to ground truth with a similarity transform fitted independently for each hand and frame. Let $(s^\star,Q^\star,b^\star)$ minimize $\sum_j\|sQp_j+b-g_j\|_2^2$, where $s\geq0$ is a uniform scale, $Q\in\mathrm{SO}(3)$ is a rotation without reflection, and $b\in\mathbb{R}^3$ is a translation. The reported error is
\begin{equation}
E_{\mathrm{PA\text{-}MPJPE}}=\frac{1}{N}\sum_{j=0}^{N-1}
\left\|s^\star Q^\star p_j+b^\star-g_j\right\|_2.
\end{equation}
The alignment minimizes squared distances, while the score averages Euclidean distances in millimeters. This metric measures the residual joint-configuration error after alignment and does not retain absolute localization or global scale error.

\enlargethispage{\baselineskip}
\textbf{2D EPE} measures agreement between the image projections of the evaluated 3D joints:
\begin{equation}
E_{\mathrm{2D}}=\frac{1}{N}\sum_{j=0}^{N-1}
\left\|\pi(P_j)-\pi(G_j)\right\|_2,
\end{equation}
where $\pi$ is the calibrated projection associated with the evaluation image and the error is in pixels. Both joint sets use the same projection within each run. This metric evaluates projections of the reconstructed 3D joints, rather than outputs of the separate 2D prediction branch. Comparisons across methods require the same image projection and resolution.

For ARCTIC, $\pi$ projects into the full-resolution $2800\times2000$
globally undistorted image using the original intrinsic matrix.

\end{document}